\documentclass{article}

\PassOptionsToPackage{numbers, compress}{natbib}

 \usepackage[preprint]{neurips_2026}

\usepackage[utf8]{inputenc} % allow utf-8 input
\usepackage[T1]{fontenc}    % use 8-bit T1 fonts
\usepackage{hyperref}       % hyperlinks
\usepackage{url}            % simple URL typesetting
\usepackage{booktabs}       % professional-quality tables
\usepackage{amsfonts}       % blackboard math symbols
\usepackage{amsmath}
\usepackage{amssymb}
\usepackage{nicefrac}       % compact symbols for 1/2, etc.
\usepackage{microtype}      % microtypography
\usepackage{xcolor}         % colors
\usepackage{graphicx}
\usepackage{caption}
\usepackage{subcaption}
\usepackage{float}
\usepackage{afterpage}

\newcommand{\Drafter}{\mathcal{D}}
\newcommand{\Target}{\mathcal{T}}
\newcommand{\ImageReward}{\mathcal{R}}

\newcommand{\score}{q}
\newcommand{\threshold}{\tau}

\hypersetup{hidelinks}

\title{Speculative Decoding for Autoregressive Video Generation}

\author{%
  Yuezhou Hu$^{*}$ \\
  University of California, Berkeley \\
  \texttt{yuezhouhu@berkeley.edu} \\
  \And
  Jintao Zhang$^{*\dag}$ \\
  University of California, Berkeley \\
  \texttt{jintaozhang@berkeley.edu} \\
}

\begin{document}

\maketitle

\def\customfootnotetext#1#2{{%
  \let\thefootnote\relax
  \footnotetext[#1]{#2}}}

\customfootnotetext{1}{\textsuperscript{*}Equal Contribution.~~~\textsuperscript{\dag}Corresponding Author.}

\begin{abstract}
Autoregressive video diffusion is emerging as a promising paradigm for
streaming video synthesis, with step distillation serving as the
primary means of accelerating inference.
Whether speculative decoding, the dominant acceleration strategy for
large language models, can be effectively adapted to autoregressive
video generation remains an open question, because video blocks are
continuous spatiotemporal tensors with no token-level distribution for
exact rejection sampling.
We introduce \texttt{SDVG}, which brings speculative decoding to
block-based autoregressive video diffusion by replacing token
verification with an image-quality router.
A 1.3B drafter proposes candidate blocks via four denoising steps;
each block is VAE-decoded and scored by ImageReward using
\emph{worst-frame aggregation}---taking the minimum per-frame reward
to catch single-frame artifacts that averaging would mask.
Blocks scoring above a fixed threshold~$\tau$ are accepted into the
14B target's KV cache; the rest are regenerated by the target.
Two additional design choices prove critical: the first block is
always force-rejected to anchor scene composition, and $\tau$ serves
as a single knob that traces a smooth quality--speed Pareto frontier.
On 1003 MovieGenVideoBench prompts ($832{\times}480$), \texttt{SDVG}
retains ${98.1\%}$ of target-only VisionReward quality
($0.0773$ vs.\ $0.0788$) at a \textbf{1.59$\times$} speedup with
$\tau{=}{-}0.7$, and reaches \textbf{2.09$\times$} at $95.7\%$
quality retention---while consistently outperforming draft-only
generation by over $+17\%$.
The framework is training-free, requires no architectural changes, and
can be seamlessly integrated into existing autoregressive video generation pipelines.
\end{abstract}

\afterpage{%
  \begin{figure}[t!]
    \centering
    \includegraphics[width=\linewidth]{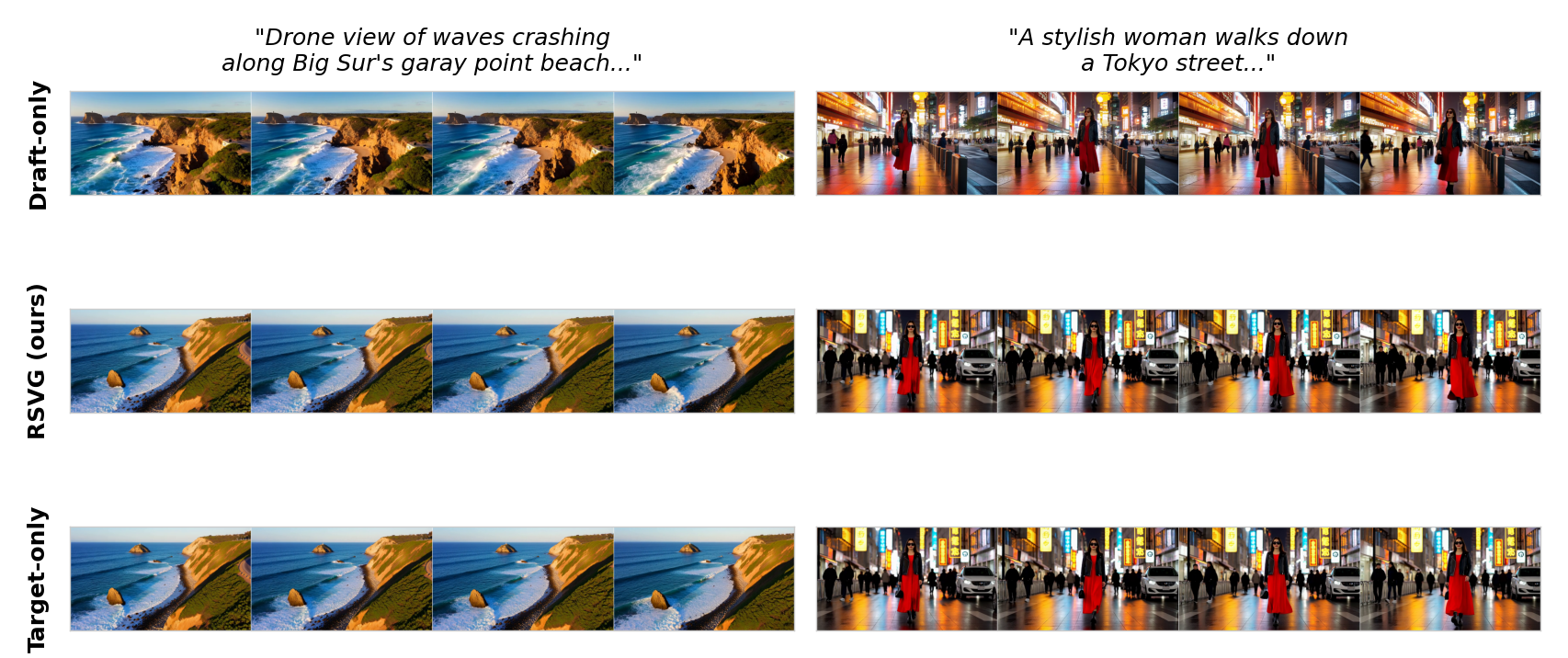}
    \caption{%
      \textbf{Qualitative comparison on MovieGenVideoBench.}
      Each row shows 4 uniformly sampled frames from a generated video.
      Draft-only (left) is fast but produces lower-fidelity output;
      Target-only (right) achieves the highest quality at the cost of
      slow inference.
      \texttt{SDVG} (center) closely matches target quality while running
      \textbf{1.59$\times$ faster}.
      Prompts: (top) ``Drone view of waves crashing against the rugged
      cliffs along Big Sur\ldots'';
      (bottom) ``A stylish woman walks down a Tokyo street\ldots''.
    }
    \label{fig:teaser}
  \end{figure}%
}

%%%%%%%%%%%%%%%%%%%%%%%%%%%%%%%%%%%%%%%%%%%%%%%%%%%%%%%%%%%%%%%%%%%%%%%%%%%%
\section{Introduction}

Autoregressive video generation has recently emerged as a compelling
paradigm for efficient, streaming video synthesis.
Unlike conventional video diffusion models that generate all frames
jointly \citep{videoworldsimulators2024,polyak2024movie}, autoregressive approaches
produce video block by block, conditioning each new block on previously
generated content through a shared key-value (KV) cache---mirroring the
autoregressive paradigm of large language models (LLMs).
This design eliminates exposure bias and enables streaming generation:
frames can be displayed as they are produced, rather than waiting for the
full sequence to complete.
Self-Forcing \citep{huang2025selfforcingbridgingtraintest} exemplifies this approach, training
a causal video diffusion transformer with self-generated conditioning that
achieves real-time video output on a single GPU.

Despite this structural efficiency advantage, state-of-the-art
autoregressive video models are built on 10B+ parameter transformers, which is still computationally demanding.
For example, frontier open-source 14B autoregressive video generation models (such as \citet{krea_realtime_14b})
require high-end GPUs (e.g. NVIDIA B200) to achieve real-time throughput.
Meanwhile, compact 1B-scale video models, such as
 \citet{wan2025wanopenadvancedlargescale}, run at less than one-quarter of the
computational cost but produce lower but
still reasonable quality.
Thus, the central question is then:

\begin{center}
\emph{Can we capture the speed of small
models while retaining the quality of large ones?}
\end{center}

Speculative decoding for LLMs \citep{leviathan2023fastinferencetransformersspeculative,chen2023acceleratinglargelanguagemodel}
offers a compelling blueprint: a small draft model proposes candidate
outputs, and the large target model is invoked only when necessary.
The block-by-block structure of autoregressive video generation is
especially well-suited to this paradigm---each generated block is a
self-contained unit that can be evaluated before being committed to the
KV cache, making per-block routing a natural design choice.

Making large and small models cooperate effectively is, however,
non-trivial.
Recent work has explored related ideas.
T-Stitch \citep{pan2024tstitchacceleratingsamplingpretrained},
SRDiffusion \citep{cheng2025srdiffusionacceleratevideodiffusion}, and
HybridStitch \citep{sun2026hybridstitch}
all propose splitting the denoising trajectory between models at the
noise-step level: the small model handles certain steps and the large
model handles others.
MoDM \citep{xia2025modmefficientservingimage} routes entire generation requests between models
at the serving system level based on a caching mechanism.
While effective in their respective settings, these approaches were not
designed for autoregressive video generation and carry meaningful
limitations.
T-Stitch, SRDiffusion, and HybridStitch use fixed step splits without detecting or
correcting poor drafts, while MoDM relies on cache hits and lacks per-block
quality guarantees. All four require extra trajectory or system-level
engineering, increasing deployment complexity.

A further challenge distinguishes video from LLMs: classical speculative
decoding accepts or rejects drafts via exact token-probability comparisons
\citep{leviathan2023fastinferencetransformersspeculative}.
Video blocks are continuous, high-dimensional spatiotemporal tensors with
no associated logit distribution, making token-level verification
inapplicable.
This leaves \emph{speculative generation for autoregressive video} as an open problem.

In this work, we propose \textbf{Speculative Decoding for Autoregressive Video
Generation} (\textbf{\texttt{SDVG}}), a training-free, plug-and-play framework
that requires no architectural changes to either the drafter or target.
For each video block, the drafter generates a candidate; an
image quality router then decides, per
block, whether to accept the draft or invoke the target for
regeneration.
Our key insight is that a plain image-quality signal, applied
block-by-block with a fixed threshold, is sufficient to match
target-only quality without any step-level trajectory engineering---simplicity
is a feature, not a limitation.
On 1003 MovieGenVideoBench prompts at $832{\times}480$ resolution, \texttt{SDVG}
achieves $98.1\%$ of target-only VisionReward quality ($0.0773$
vs.\ $0.0788$) at a $\mathbf{1.59\times}$ speedup.
Importantly, \texttt{SDVG} is \emph{orthogonal} to step-level methods such as
T-Stitch or SRDiffusion, which can be directly applied to the target
model's generation steps within \texttt{SDVG}, providing a composable path to
further speedup.
Our main contributions are as follows:
\begin{itemize}
  \item We propose \texttt{SDVG}, a training-free speculative generation framework
    for autoregressive video diffusion that routes each block between a
    drafter and a target based on image quality routing,
    achieving a $1.59\times$ speedup with $98.1\%$ quality retention.
  \item We identify three video-specific design choices critical to making
    reward-guided routing effective: a fixed
    ImageReward threshold that provides a simple, calibration-free
    quality--speed knob, mandatory first-block regeneration to
    anchor scene composition, and worst-frame quality scoring to surface
    single-frame artifacts masked by block averages.
  \item We demonstrate that without any complex
    step-level trajectory engineering, a plain reward routing signal
    suffices to match large-model quality---establishing a new,
    simpler baseline for collaborative video generation.
\end{itemize}

%%%%%%%%%%%%%%%%%%%%%%%%%%%%%%%%%%%%%%%%%%%%%%%%%%%%%%%%%%%%%%%%%%%%%%%%%%%%
\begin{figure}[!t]
  \centering
  \includegraphics[width=\linewidth]{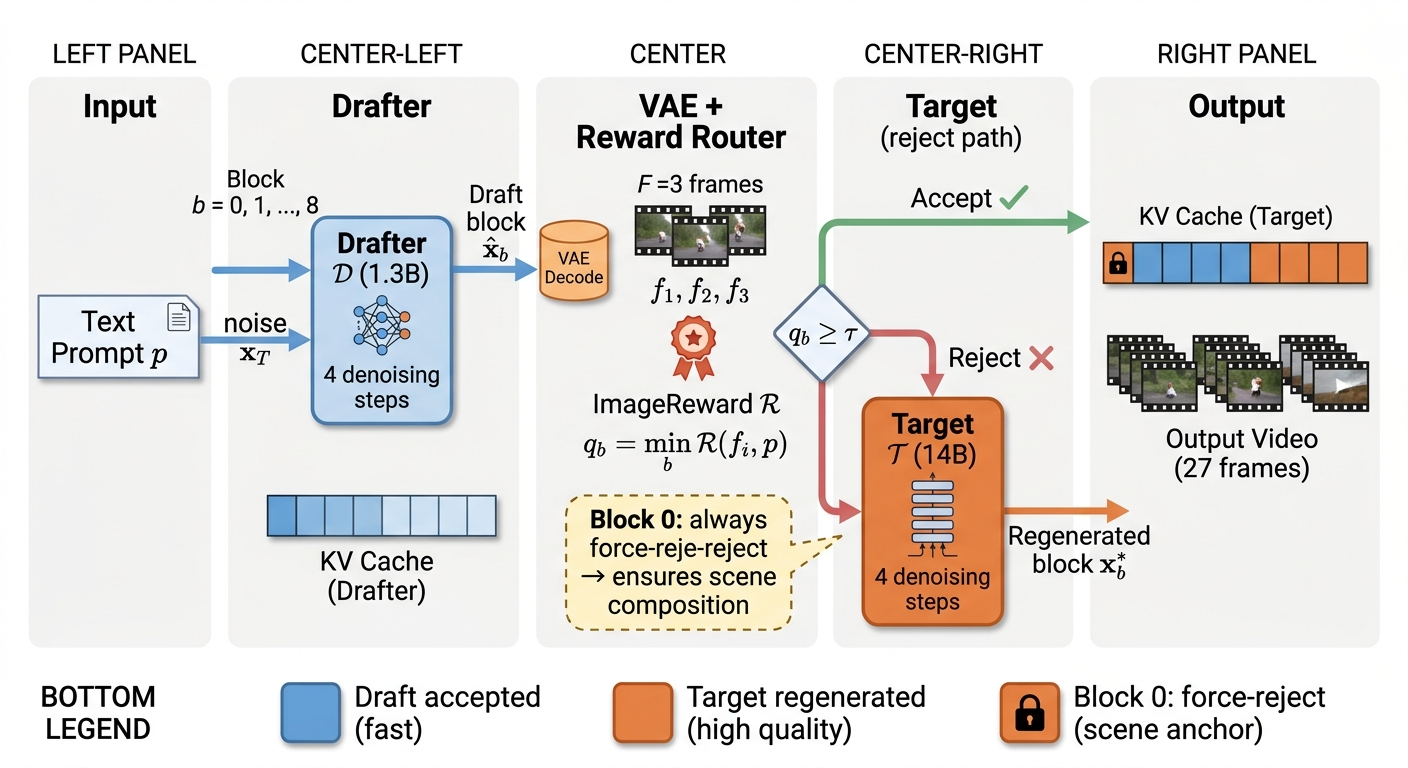}
  \caption{%
    \textbf{\texttt{SDVG} inference pipeline.}
    For each video block, the 1.3B drafter generates a candidate in 4
    denoising steps.
    Block~0 is always force-rejected to anchor scene composition.
    For blocks 1--8, the draft is VAE-decoded and scored by ImageReward
    using min-frame aggregation ($q_b = \min_i \mathcal{R}(f_i, p)$).
    If $q_b \geq \tau$, the draft is accepted (blue) and committed to the
    target KV cache; otherwise the 14B target regenerates the block
    (orange).
  }
  \label{fig:pipeline}
\end{figure}

\section{Background}

\paragraph{Video generation.}
Diffusion-based video models have advanced from pixel-space approaches
\citep{ho2022videodiffusionmodels} to large latent transformer architectures
\citep{videoworldsimulators2024,polyak2024movie}.
Inference efficiency has been improved primarily through step distillation
\citep{yin2024onestepdiffusiondistributionmatching,zhang2025turbodiffusion,wang2024phased} and GPU kernel optimization~\citep{zhang2025sageattention,zhang2024sageattention2,zhang2025sageattention2++,zhang2025sageattention3,zhangspargeattention,zhang2025sla}. They are all orthogonal to \texttt{SDVG}.

% , and
% distributed parallelism \citep{li2024distrifusiondistributedparallelinference}
% flow-matching \citep{lipman2023flowmatchinggenerativemodeling},

\paragraph{Autoregressive video generation.}
Autoregressive video generation models future video blocks causally based on previously generated content. Early training methods rely on ground-truth history during training, resulting in a mismatch with autoregressive inference and consequently causing exposure bias. More recent methods, such as Self-Forcing \citep{huang2025selfforcingbridgingtraintest}, aim to improve training–inference consistency in block-wise video generation.

\paragraph{Hierarchical video generation.}
T-Stitch \citep{pan2024tstitchacceleratingsamplingpretrained},
SRDiffusion \citep{cheng2025srdiffusionacceleratevideodiffusion}, and
HybridStitch \citep{sun2026hybridstitch}
split the denoising trajectory or space between a small and a large model at fixed
noise levels, achieving training-free acceleration.
MoDM \citep{xia2025modmefficientservingimage} routes entire requests to smaller models on cache
hits, reducing average serving time by $2.5\times$.
However, these methods apply content-agnostic step-level splits and were not
designed for block-level autoregressive video.

\paragraph{Speculative decoding.}
Speculative decoding \citep{leviathan2023fastinferencetransformersspeculative,chen2023acceleratinglargelanguagemodel}
pairs a small drafter with a large target: the drafter proposes tokens
and the target verifies them in one pass, preserving the target
distribution exactly.
RSD \citep{liao2025rewardguidedspeculativedecodingefficient} lifts acceptance to the reasoning-step level using a
Process Reward Model (PRM), directly inspiring \texttt{SDVG}.
The key difference is that video blocks are continuous tensors with no
token distribution, so exact rejection sampling is inapplicable; we
replace the PRM with an image quality model as a block-level proxy.

%%%%%%%%%%%%%%%%%%%%%%%%%%%%%%%%%%%%%%%%%%%%%%%%%%%%%%%%%%%%%%%%%%%%%%%%%%%%
\section{Method: \texttt{SDVG}}
\label{sec:method}

Given text prompt $p$, drafter $\Drafter$ and target $\Target$,
we seek a routing policy $\pi: \mathbb{R} \to \{0,1\}$ that maps a
per-block quality score $\score_b \in \mathbb{R}$ to accept (1) or
reject (0), optimizing the trade-off between video quality and inference
speed.

\textbf{Inference flow.}
For each block $b$, $\Drafter$ runs $S$ denoising steps to produce a
candidate $\hat{\mathbf{x}}_b$.
The drafter KV cache $\mathbf{K}^{\Drafter}$ is updated unconditionally,
ensuring $\Drafter$ always conditions on its own prior outputs.
The candidate is decoded by the VAE and scored by the router.
If $\score_b \geq \threshold$, the draft is \emph{accepted}:
$\hat{\mathbf{x}}_b$ is committed to the target's KV cache
$\mathbf{K}^{\Target}$ and the decoded frames are emitted directly.
If \emph{rejected}, $\Target$ runs $S$ denoising steps from the same
initial noise to produce $\mathbf{x}^*_b$, updating $\mathbf{K}^{\Target}$.
The VAE decode cache is cloned before draft scoring and restored on
rejection to prevent temporal inconsistency across blocks.
In \texttt{SDVG} the threshold $\threshold$ is a fixed scalar calibrated
offline.

\textbf{Worst-frame aggregation.}
The block quality score is the \emph{minimum} per-frame reward over the
$F$ decoded frames:
\begin{equation}
  \score_b = \min_{i=1}^{F}
  \ImageReward\!\left(\mathbf{f}^{(b)}_i,\, p\right).
  \label{eq:minscore}
\end{equation}
Here $\ImageReward(\mathbf{f}, p)$ denotes the reward of a single decoded
frame $\mathbf{f}$ given prompt $p$.
Using the minimum rather than the mean catches blocks with one severely
degraded frame---a visual artifact that average scoring would mask.

\textbf{Force-reject the first block.}
Block $b=0$ is always regenerated by $\Target$, regardless of its draft
score.
Block 0 lacks any KV context from prior blocks and establishes the scene
composition, foreground subjects, and visual style that all subsequent
blocks inherit through the shared cache.
Accepting a draft at this position risks propagating irreversible layout
errors throughout the video.

%%%%%%%%%%%%%%%%%%%%%%%%%%%%%%%%%%%%%%%%%%%%%%%%%%%%%%%%%%%%%%%%%%%%%%%%%%%%
\section{Experiments}
\label{sec:experiments}

\subsection{Experimental Setup}
\label{sec:setup}

\paragraph{Models.}
We evaluate \texttt{SDVG} on a pair of autoregressive video diffusion models built
on the Wan2.1 architecture \citep{wan2025wanopenadvancedlargescale}.
The \emph{target model} is Krea Realtime Video 14B \citep{krea_realtime_14b},
distilled from Wan2.1-T2V-14B via Self-Forcing \citep{huang2025selfforcingbridgingtraintest}.
The \emph{drafter} is the original Wan2.1-T2V-1.3B Self-Forcing model.
Both models share the same causal attention backbone with KV caching via
RoPE positional embeddings, and run 4 denoising steps per block using the
schedule $\mathbf{t} = [1000, 937, 833, 625, 0]$ in bfloat16 precision
(guidance scale 3.0, timestep shift 5.0).
The reward router uses ImageReward \citep{xu2023imagerewardlearningevaluatinghuman}, an
off-the-shelf text-image reward model, to score decoded draft frames.

\paragraph{Generation protocol.}
Each video consists of $B{=}9$ autoregressive blocks. Each block corresponds to 3 latent frames (27 latent frames in total). The causal VAE
decoder produce $9$ pixel frames for the first video block and $12$ pixel frames per later video block at $832 \times 480$ resolution.
All runs use a fixed random seed (42) for reproducibility.
The routing threshold is set to $\tau{=}{-}0.7$ (min-frame ImageReward)
unless otherwise noted.

\paragraph{Hardware and implementation.}
All experiments are conducted on two NVIDIA RTX A6000 GPUs (48\,GB each).
GPU\,0 hosts the diffusion transformer (both target and drafter);
GPU\,1 hosts the text encoder (UMT5-XXL), causal VAE, and ImageReward.
CUDA streams overlap cross-device transfers with compute so that reward
scoring does not block denoising.
The VAE decode cache is cloned before draft scoring and restored upon
rejection to preserve temporal consistency across blocks.

\paragraph{Evaluation benchmarks and metrics.}
We draw prompts from MovieGenVideoBench \citep{polyak2024movie}, which
spans diverse categories including landscapes, animals, human activities,
and cinematic footage.
We report results on the full 1003-prompt set.
Video quality is measured by {VisionReward} \citep{xu2026visionrewardfinegrainedmultidimensionalhuman},
a VQA-based metric that aggregates 29 questions covering visual quality,
temporal consistency, motion naturalness, and text--video alignment.
Efficiency is measured by {wall-clock time} per video (excluding
model loading and warmup), and the resulting {speedup} relative to
target-only generation.

\paragraph{Baselines.}
We compare against two boundary baselines:
\emph{Draft-only}---all blocks generated by the 1.3B drafter
(maximum speed, lowest quality);
\emph{Target-only}---all blocks generated by the 14B target
(minimum speed, highest quality).
\texttt{SDVG} operates between these extremes by selectively routing blocks.

\subsection{Main Results}
\label{sec:main-results}

Table~\ref{tab:main} presents \texttt{SDVG} results across a sweep of fixed
min-frame thresholds on 1003 MovieGenVideoBench prompts, alongside the
two boundary baselines.

\begin{table}[t]
  \caption{%
    Main results on 1003 MovieGenVideoBench prompts
    ($832{\times}480$, 9 blocks/video).
    VR = VisionReward (higher is better).
    Time = average wall-clock time per video.
    Speedup is relative to target-only.
    Accept rate excludes block~0 (always force-rejected).
  }
  \label{tab:main}
  \centering
  \setlength{\tabcolsep}{13pt}
  % \small
  \begin{tabular}{lccccc}
    \toprule
    Method & VR $\uparrow$ & Time (s) & Speedup & Accept \\
    \midrule
    Target-only                       & 0.0788 & 97.0 & 1.00$\times$ & ---    \\
    \midrule
    \texttt{SDVG} ($\tau{=}{-}0.7$)            & 0.0773 & 60.9 & 1.59$\times$ & 73.1\% \\
    \texttt{SDVG} ($\tau{=}{-}0.8$)            & 0.0769 & 58.6 & 1.66$\times$ & 74.9\% \\
    \texttt{SDVG} ($\tau{=}{-}0.9$)            & 0.0771 & 58.3 & 1.66$\times$ & 76.4\% \\
    \texttt{SDVG} ($\tau{=}{-}1.0$)            & 0.0764 & 57.2 & 1.69$\times$ & 78.0\% \\
    \texttt{SDVG} ($\tau{=}{-}1.5$)            & 0.0757 & 51.6 & 1.88$\times$ & 83.4\% \\
    \texttt{SDVG} ($\tau{=}{-}2.0$)            & 0.0756 & 47.4 & 2.05$\times$ & 87.5\% \\
    \texttt{SDVG} ($\tau{=}{-}2.5$)            & 0.0754 & 46.4 & 2.09$\times$ & 88.9\% \\
    \midrule
    Draft-only                        & 0.0644 & 25.7 & 3.77$\times$ & ---    \\
    \bottomrule
  \end{tabular}
\end{table}

\begin{figure}[t]
  \centering
  \includegraphics[width=0.7\linewidth]{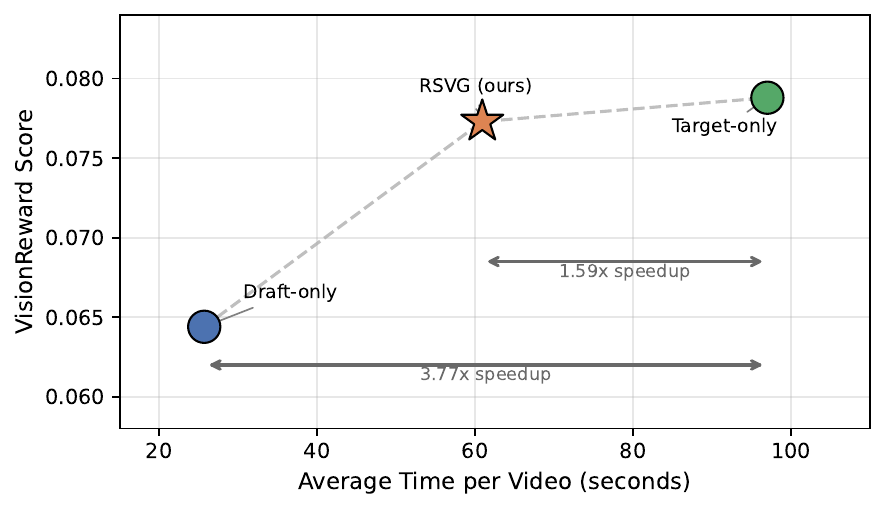}
  \caption{%
    \textbf{Quality--speed Pareto curve.}
    \texttt{SDVG} ($\tau{=}{-}0.7$) achieves 98.1\% of target quality at
    1.59$\times$ speedup, operating between draft-only and target-only.
  }
  \label{fig:pareto}
\end{figure}

\paragraph{Quality--time tradeoff.}
By sweeping the threshold $\tau$ from $-0.7$ to $-2.5$, \texttt{SDVG} traces a
smooth Pareto frontier between the two baselines
(Figure~\ref{fig:pareto}).
At the conservative end ($\tau{=}{-}0.7$), \texttt{SDVG} retains
${98.10\%}$ of target-only VisionReward (0.0773 vs.\ 0.0788)
with a \textbf{1.59$\times$} speedup.
As the threshold relaxes, inference time continues to drop:
$\tau{=}{-}1.0$ pushes the speedup to \textbf{1.69$\times$} while
preserving 96.95\% of target quality.
At the aggressive end ($\tau{=}{-}2.5$), \texttt{SDVG} reaches
\textbf{2.09$\times$} speedup at 95.69\% quality retention---still
substantially above draft-only (0.0754 vs.\ 0.0644, a $+17.1\%$ gap).
The diminishing returns beyond $\tau{=}{-}1.5$ suggest that most
quality-critical blocks already have ImageReward scores above $-1.5$;
further relaxation buys little additional speed.

\paragraph{Quality--acceptance rate tradeoff.}
An alternative view of the same frontier is the relationship between
VisionReward and draft acceptance rate.
As the accept rate rises from 73.1\% ($\tau{=}{-}0.7$) to 78.0\%
($\tau{=}{-}1.0$), VisionReward decreases only marginally
(0.0773 $\to$ 0.0764, $-0.12\%$ absolute).
This near-flat region indicates that the additional drafts admitted by
relaxing $\tau$ are borderline cases whose quality is close to the
target model's output---the reward-guided router selectively accepts
drafts that would not degrade perceptual quality.
Beyond 78\% acceptance, quality begins to drop more noticeably
(0.0757 at 83.4\%, 0.0754 at 88.9\%), as increasingly low-scoring
drafts are admitted.
The inflection around $\tau \in [-1.0, -0.7]$ (73--78\% acceptance)
identifies the operating regime where \texttt{SDVG} delivers the most favorable
quality--efficiency balance.

\subsection{Ablation Studies}
\label{sec:ablation}

We ablate the two key design choices in \texttt{SDVG}---the scoring strategy and
the routing signal---to isolate their contributions.
Table~\ref{tab:ablation} reports results on 1003 MovieGenVideoBench
prompts.
All ablations use the same force-reject-block-0 policy and fixed
thresholds for a controlled comparison.

\paragraph{Reward-guided vs.\ random routing.}
To verify that the ImageReward signal is load-bearing, we replace the
reward router with random accept/reject decisions (at a matched overall
accept rate).
Random routing yields VisionReward 0.0706, a sharp drop from
the reward-guided \texttt{SDVG} (0.0773 at $\tau{=}{-}0.7$) and even below the
force-reject-block-0 baseline (0.0757 from the main table).
Without a quality signal, the router accepts artifact-heavy blocks and
rejects clean ones with equal probability, negating the benefit of
selective regeneration.

\paragraph{Min-frame vs.\ average-frame scoring.}
Our default \texttt{SDVG} uses min-frame aggregation (Eq.~\ref{eq:minscore}),
which flags blocks where even a single frame is degraded.
Replacing it with average-frame scoring consistently underperforms:
at comparable accept rates, the avg-frame variants achieve lower
VisionReward (e.g., 0.0755 at 78.4\% acceptance vs.\ 0.0773 at 73.1\%
for min-frame $\tau{=}{-}0.7$).
This confirms that averaging masks per-frame artifacts---a single
corrupted frame among $F{=}3$ produces visible temporal flickering that
the mean score fails to catch, leading the router to accept low-quality
blocks.

\begin{table}[t]
  \caption{%
    \textbf{Ablation study.}
    1003 MovieGenVideoBench prompts, $832{\times}480$, 9 blocks/video.
    VR = VisionReward (higher is better).
    Time = average wall-clock time per video.
    Accept rate excludes block~0 (always force-rejected).
    ``---'' indicates not applicable.
  }
  \label{tab:ablation}
  \centering
  \setlength{\tabcolsep}{13pt}
  \begin{tabular}{lcccc}
    \toprule
    Method & VR $\uparrow$ & Time (s) & Speedup & Accept \\
    \midrule
    Target-only                       & 0.0788 & 97.0 & 1.00$\times$ & ---    \\
    \midrule
    \textbf{\texttt{SDVG} ($\tau{=}{-}0.7$, ours)} & \textbf{0.0773} & 60.9 & 1.59$\times$ & 73.1\% \\
    \midrule
    \multicolumn{5}{l}{\textit{Avg-frame scoring (replacing min-frame)}} \\
    Avg-frame $\tau{=}{-}0.2$         & 0.0767 & 63.2 & 1.54$\times$ & 70.2\% \\
    Avg-frame $\tau{=}{-}0.5$         & 0.0759 & 59.0 & 1.64$\times$ & 75.3\% \\
    Avg-frame $\tau{=}{-}0.7$         & 0.0755 & 56.9 & 1.71$\times$ & 78.4\% \\
    \midrule
    \multicolumn{5}{l}{\textit{Routing signal ablation}} \\
    Force-reject $+$ random routing   & 0.0771 & 58.2 & 1.67$\times$ & 70.3\% \\
    Random routing                    & 0.0706 & 60.2  & 1.61$\times$ & 70.0\% \\
    \midrule
    Draft-only                        & 0.0644 & 25.7 & 3.77$\times$ & ---    \\
    \bottomrule
  \end{tabular}
\end{table}

%%%%%%%%%%%%%%%%%%%%%%%%%%%%%%%%%%%%%%%%%%%%%%%%%%%%%%%%%%%%%%%%%%%%%%%%%%%%
\section{Limitations}

\textbf{Distributional bias.}
Unlike exact-rejection LLM speculative decoding, \texttt{SDVG} accepts a
distributional shift toward the drafter.
A lower accept rate (stricter threshold) reduces the gap at the cost of speedup.

\textbf{ImageReward as a proxy.}
ImageReward was trained on text-image pairs and evaluates frames
independently, missing temporal consistency and motion quality.
A dedicated video-block quality model would improve the routing signal.

\textbf{Wasted draft computation.}
For rejected blocks (including forced block-0 rejections),
the drafter forward pass and VAE decode are wasted computation.
Batching or speculative VAE decoding could reduce this overhead.

%%%%%%%%%%%%%%%%%%%%%%%%%%%%%%%%%%%%%%%%%%%%%%%%%%%%%%%%%%%%%%%%%%%%%%%%%%%%
\section{Conclusion}

We presented \texttt{SDVG}, a reward-guided speculative video generation framework
for autoregressive video diffusion that achieves 98.1\% of target quality
at 1.59$\times$ speedup.
The key design choices---forced first-block regeneration and worst-frame
quality scoring---address the specific challenges of block-level video
speculative decoding.
A single fixed threshold provides a simple quality--speed knob: sweeping
it from $-0.7$ to $-2.5$ traces a smooth Pareto frontier reaching up to
$2.09\times$ speedup while remaining well above draft-only quality.
\texttt{SDVG} can be applied to any Self-Forcing-style autoregressive video
model with a drafter-target pair, opening the door to reward-guided
inference-time compute allocation for video generation.

% \begin{ack}
% Use unnumbered first level headings for the acknowledgments. All acknowledgments
% Do {\bf not} include this section in the anonymized submission, only in the final paper. You can use the \texttt{ack} environment provided in the style file to automatically hide this section in the anonymized submission.
% \end{ack}

\bibliographystyle{abbrvnat}
\bibliography{references}

%%%%%%%%%%%%%%%%%%%%%%%%%%%%%%%%%%%%%%%%%%%%%%%%%%%%%%%%%%%%%%%%%%%%%%%%%%%%

% \appendix
%
% \section{Technical Appendix}
%
% Additional results, figures, and analysis may be placed here. There is no
% page limit for appendices, but reviewers are not required to read them.

%%%%%%%%%%%%%%%%%%%%%%%%%%%%%%%%%%%%%%%%%%%%%%%%%%%%%%%%%%%%%%%%%%%%%%%%%%%%

% \newpage
% \input{checklist.tex}

\end{document}